\documentclass[journal]{journal}
\ifCLASSINFOpdf
   \usepackage[pdftex]{graphicx}
\else
\fi
\usepackage[tight,footnotesize]{subfigure}
\usepackage{url}

\usepackage{hyperref}

\begin{document}
%
\title{Quantifying urban streetscapes with deep learning: focus on aesthetic evaluation}
%
%
%

\author{Yusuke~Kumakoshi*,
        Shigeaki~Onoda,
        Tetsuya~Takahashi,
        Yuji~Yoshimura
\thanks{Y. Kumakoshi (*corresponding author) and Y. Yoshimura are with Research Center for Advanced Science and Technology, the University of Tokyo, Japan. e-mail: ykuma@cd.t.u-tokyo.ac.jp, yyyoshimura@cd.t.u-tokyo.ac.jp}
\thanks{S. Onoda is with Softbank Corp., Tokyo, Japan. e-mail: ono.fda03@gmail.com}
\thanks{T. Takahashi is with the Department of Urban Engineering, the University of Tokyo, Japan. e-mail: pknmdmkodos@gmail.com}
}

%
%

\markboth{Journal of \LaTeX\ Class Files,~Vol.~6, No.~1, January~2007}%
{Shell \MakeLowercase{\textit{et al.}}: Bare Demo of IEEEtran.cls for Journals}
%



\maketitle
\thispagestyle{empty}

\begin{abstract}
The disorder of urban streetscapes would negatively affect people’s perception of their aesthetic quality. The presence of billboards on building facades has been regarded as an important factor of the disorder, but its quantification methodology has not yet been developed in a scalable manner. To fill the gap, this paper reports the performance of our deep learning model on a unique data set prepared in Tokyo to recognize the areas covered by facades and billboards in streetscapes, respectively. The model achieved 63.17 \% of accuracy, measured by Intersection-over-Union (IoU), thus enabling researchers and practitioners to obtain insights on urban streetscape design by combining data of people’s preferences.
\end{abstract}


%
\IEEEpeerreviewmaketitle

\section{Introduction}
%
%
%
%
\IEEEPARstart{T}{he} aesthetic quality of urban streetscapes is one dimension of people’s perception of their surroundings  \cite{carp1976dimensions} and it affects the attractiveness of urban spaces \cite{weber2008aesthetics}. Among several fundamental properties of streetscapes that affect the perception, order and complexity are known as the most influential factors \cite{birkhoff2013aesthetic}, regardless of other individual factors such as experience and values \cite{portella2016visual}. However, in today’s streetscapes in city centers, both elements seem not to be satisfied because of the abundance of various signs, billboards, and eccentrically designed buildings \cite{portella2016visual}. Portella (2016) called such disorder of the streetscape as visual pollution and attributed a source of streetscapes’ disorder to “conflict between the design of commercial signs and aesthetic composition of building facades \cite{portella2016visual}.”  

While Portella (2016) focused on the presence of billboards in relation to the superficial area of building facades \cite{portella2016visual}, Ashihara (1983) focused on the perspective of people on the street: he defined the outline formed by facades or exterior walls of buildings as the primary contour, and that by billboards and signs that were installed on the exterior walls as the secondary contour \cite{ashihara1983aesthetic}. Using these notions, he argued that streetscapes perceived as beautiful tend to have more area surrounded by the primary contour than by the secondary contour. These notions reflect people’s experience on the streets, and notably, they have been used to advocate aesthetic superiority of Occidental streetscapes compared to Oriental ones (typically the Japanese ones); they have been referred to as a rationale of urban landscape design \cite{ashihara1983aesthetic}.

Despite the presence of these theoretical frameworks, there are limited attempts to evaluate the aesthetics of urban streetscapes in a quantitative manner using real data sets; therefore, the validity of the theories has not been thoroughly tested. Ahmed et al. (2019) developed a deep learning model that recognizes four categories of objects that cause visual pollutions, one of which is the presence of billboards and signs \cite{ahmed2019solving}. However, their model focused on the existence of the object and did not consider the quantity of each object in a streetscape.

One reason for the absence of empirical validation is the lack of (1) efficient methodologies to estimate the areas in streetscapes surrounded by the contours and (2) the data set of people’s perception and evaluation of streetscapes. Regarding the former, the areas may be manually calculated, but it is highly time-consuming and difficult to prepare a large amount of data. A similar difficulty of data collection applies to the latter.

To overcome the first half of the difficulties, the present study developed a deep-learning-based model on the data set composed of images collected and annotated by the authors and collaborators. Notably, it aims to operationalize the theory of the primary and secondary contour in the aesthetics of streetscapes \cite{ashihara1983aesthetic}. The scope of the present paper is limited to the development of the semantic segmentation model: further investigation on the human’s perception is an object of a future study.

The major contributions of this study are the following:
\begin{enumerate}
    \item Elaborated a unique annotation data set of billboards in Japanese urban streetscapes.
    \item Developed a semantic segmentation model of billboards, with 63.17\% of IoU.
\end{enumerate}

\section{Data set and methodology}
\subsection{Data set}
Two data sets were used in this research: Cityscapes for the primary contour and original one for the secondary contour. Cityscapes data set contains urban landscapes on German streets \cite{cordts2016cityscapes}, and is expected to serve as a training data set to extract facades and walls of buildings (the primary contour). 

Regarding the secondary contour, billboards in the Japanese urban streetscapes are expected to be different, in linguistic terms, from those contained in most of the available data sets of annotated images for semantic segmentation, because they use images taken outside Japan. Hence, to improve the prediction accuracy of billboards, we prepared a unique annotated data set. 

The original images for the data set were produced in the following manner: we recorded movies with a GoPro camera while running in the streets in Tokyo on taxis and bicycles, and then extracted images from them. First, we mounted the GoPro camera on the windshield of the taxi and recorded movies of streets from different lanes and directions so that the eventual images in the movies have variations of angles and subjects. Second, to increase the variety of perspectives in the movies, we additionally recorded movies from a bicycle with the GoPro camera mounted on the front bike basket (80 cm from the ground). The recording was done in major commercial centers in Tokyo, namely Ginza, Marunouchi, Akihabara, Shinjuku, Shibuya, Ikebukuro and Kichijoji. These sites are ideal for collecting images because some districts in these sites have numerous outdoor advertising billboards.

After the recording, we extracted images from the movies every one second-intervals, or 60 frames (e.g., the movies were recorded in 60 fps). Further, we manually excluded images that are not likely to contribute to the improvement of the model accuracy such as those containing a strong halation effect or those without any billboard objects.

For the rest of the images, the billboards in the images were annotated. We assumed that the secondary contour includes road signs and signs on glasses on building facades, in addition to billboards, because all these objects cover building facades. Further, tiny objects near the vanishing points in images were not annotated, because people on the street would not clearly recognize billboards far from them. The total amount of the annotated images was 5495, and available online\footnote{\href{https://www.kaggle.com/yusukekumakoshi/billboard-in-japanese-streetscapes}{https://www.kaggle.com/yusukekumakoshi/billboard-in-japanese-streetscapes}.}.
The data are separated for training, test, and validation with the proportion of 0.68:0.12:0.2 in the case of the final model (DeepLabV3+ downscale model, explained in the next section).
 
Further, the data were resized to be loaded on the GPU memory (NVIDIA Tesla V100 on AWS) and augmented by applying the following processes: horizontal flip, Affine transformation (shift, scale, and rotate), random crop, random contrast, and random four-point perspective transformation (IAA perspective). Only for the DeepLabV3+ downscale model, the resolution of the data was lowered: downscaled. This method was applied because the downscale in semantic segmentation is expected to improve inference on larger and wider regions in images \cite{tao2020hierarchical}, and we expect that it allows the model to recognize the billboards with enhanced accuracy.

\subsection{Model development}
We adopted a supervised approach for model development because the appearance of building facades and billboards largely depends on the local context; it would be difficult to detect facades or billboards simply from the colors of a pixel in the image (e.g., \cite{cai2018treepedia,cai2019quantifying}).

For the primary contour, we used the pre-trained model DeepLab+Xc7  \cite{chen2018encoder,alves2020low}. Regarding the secondary contour, we developed a DCNN that detects billboards with better precision than the above model, because the first model’s inference on billboards was not precise enough.

We used DeepLabV3+ \cite{chen2018encoder}’s architecture to train a semantic segmentation model for billboard segmentation. First, ResNet50 architecture \cite{he2016deep} and ImageNet data set \cite{deng2009imagenet} were used to pre-train the encoder (Benchmark model). This model was trained in the early stage of this study; therefore, the amount of data used for the training was limited. Second, we trained a DeepLabV3+’s architecture that includes the same encoder on our unique data set (DeepLabV3+ downscale model). The data used to train this second model is slightly different from that for the benchmark model in terms of quantity (e.g., the amount was richer for the second model) and data augmentation (e.g., downscale and random contrast were added in the augmentation process for the second model). 

The models for the secondary contour were trained using Pytorch framework \cite{paszke2019pytorch} on NVIDIA Tesla V100 on AWS.

\subsection{Metrics of model precision}
As an evaluation metric for the accuracy of billboard segmentation, mean Intersection-over-Union (IoU) was employed: we calculated the overlapping area of the two layers, the annotated images, and the predicted images produced by the model. This metric has been widely used to evaluate the performance of semantic segmentation \cite{cordts2016cityscapes,rezatofighi2019generalized}.

\subsection{Integrated analysis}
The areas covered by the primary contour and by the secondary contours in each image were calculated on a pixel basis, to characterize the aesthetic aspect of urban streetscapes. Originally, Ashihara (1983) focused on the proportions of the areas covered by the contours over the whole image of the streetscape. Because these two quantities have the same denominator (the area of the whole image), it is sufficient to calculate the ratio of the areas covered by the two contours. Given that each pixel in an image is classified into (1) the area covered by the primary contour, (2) the area covered by the secondary contour, (3) sky, or (4) road, the same argument applies to a case in which we consider the sky and the road: the denominator is the area that is neither sky nor road for both cases.

One drawback of using the quantity is the sensitivity to the viewpoint: two streetscape images taken at the same place but with different heading angles yield different estimations. In other words, it characterizes a streetscape from a fixed viewpoint, not considering the entire surrounding of the place. However, this reflects people’s visual experiences in that their perception of the streetscape depends on their relative location and perspective to the surrounding building. Rather, more attention should be paid to the comparison of different images by keeping the same perspective.

Several images taken in urbanized areas in Tokyo were inferred using the developed model, as a case study.

\section{Results}
\subsection{Accuracy of the models}
Table \ref{tab:accuracy} shows the accuracy of billboard segmentation (the secondary contour) for each model. The benchmark model was trained using 439 images, validated with 148 images, and tested with 147 images (0.6:0.2:0.2). DeepLabV3+ downscale model was trained using 3736 images, validated with 659 images, and tested with 1099 images (0.68:0.12:0.2). In terms of Mean IoU on the test sets, DeepLabV3+ downscale model was more than twice as accurate as the benchmark model.

\begin{table}[h]
\centering
\begin{tabular}{c|ccc}
    & Mean IoU & \# epoch & \# trained images\\ \hline
Benchmark & 23.74\%  &       40    &   439      \\
DeepLabV3+ downscale &  63.17\%  &     40   & 3736   
\end{tabular}
\caption{Accuracy comparison between models.}
\label{tab:accuracy}
\end{table}

\subsection{Case study in Tokyo}
\begin{figure*}[h]
    \centering
    \begin{tabular}{ccc}
      \begin{minipage}[t]{0.3\linewidth}
        \centering
            \includegraphics[keepaspectratio,scale=0.039]{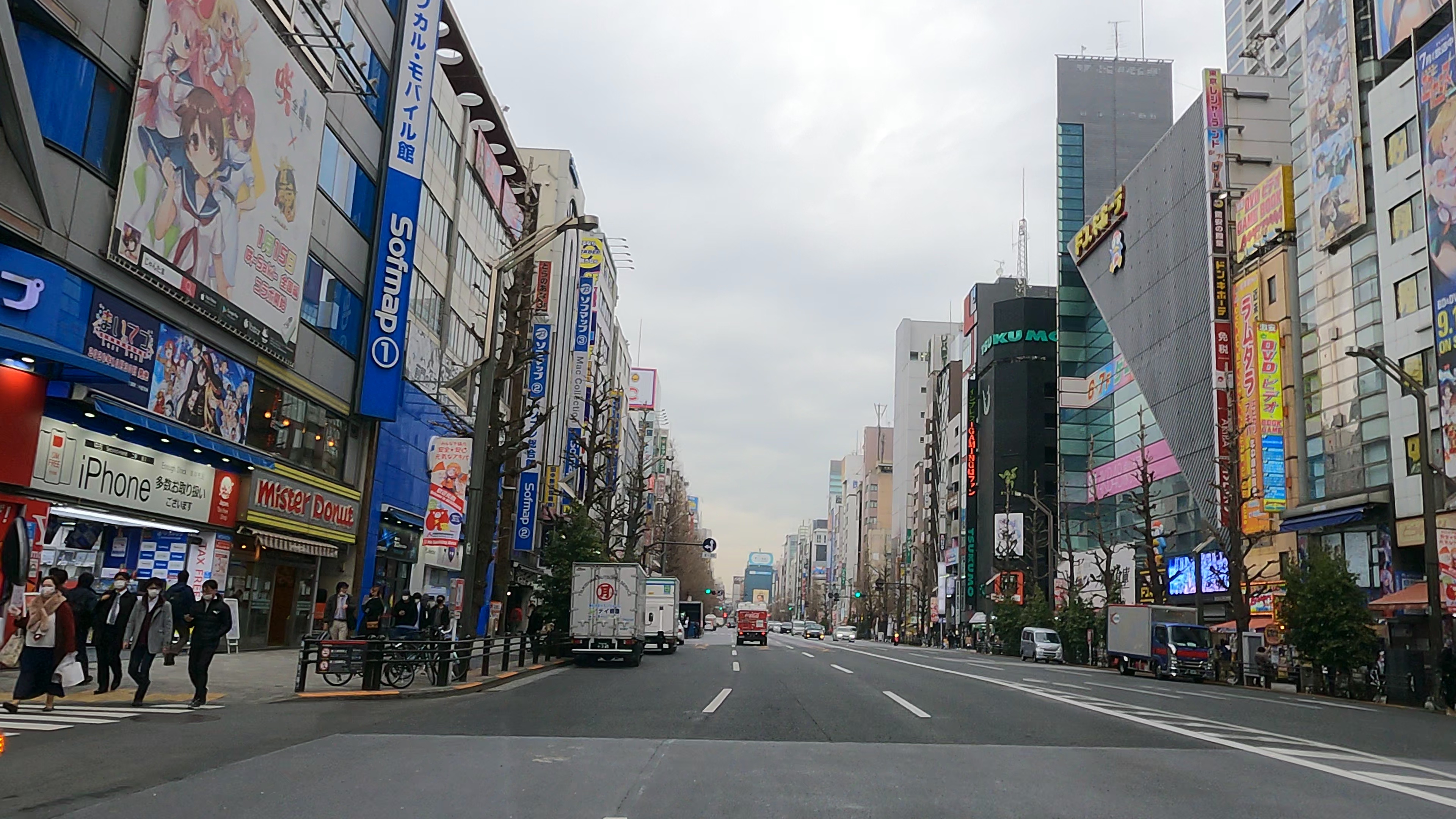}
      \end{minipage} &
      
        \begin{minipage}[t]{0.3\linewidth}
        \centering
            \includegraphics[keepaspectratio,scale=0.078]{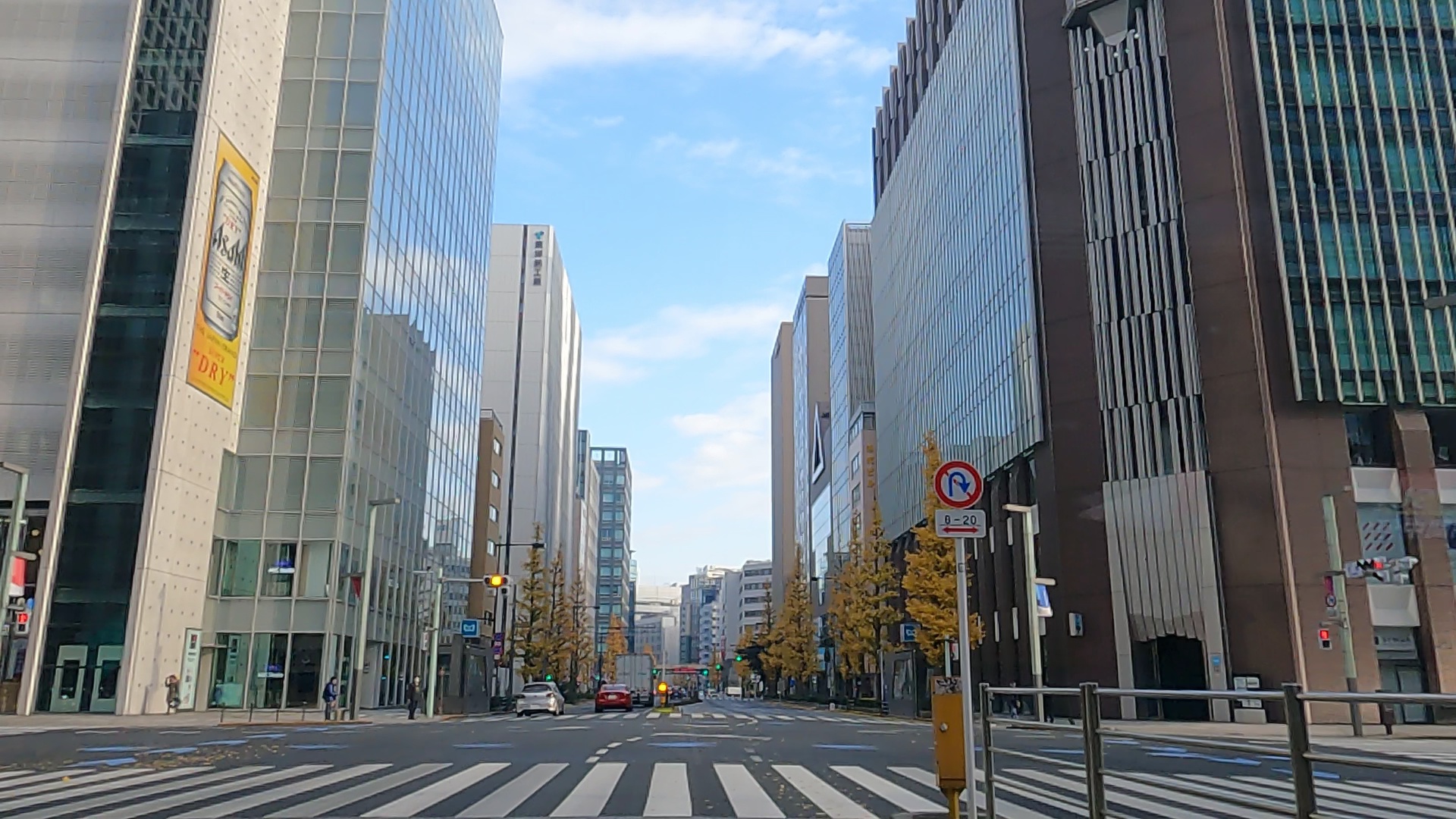}
      \end{minipage} &
      
    \begin{minipage}[t]{0.3\linewidth}
        \centering
            \includegraphics[keepaspectratio,scale=0.039]{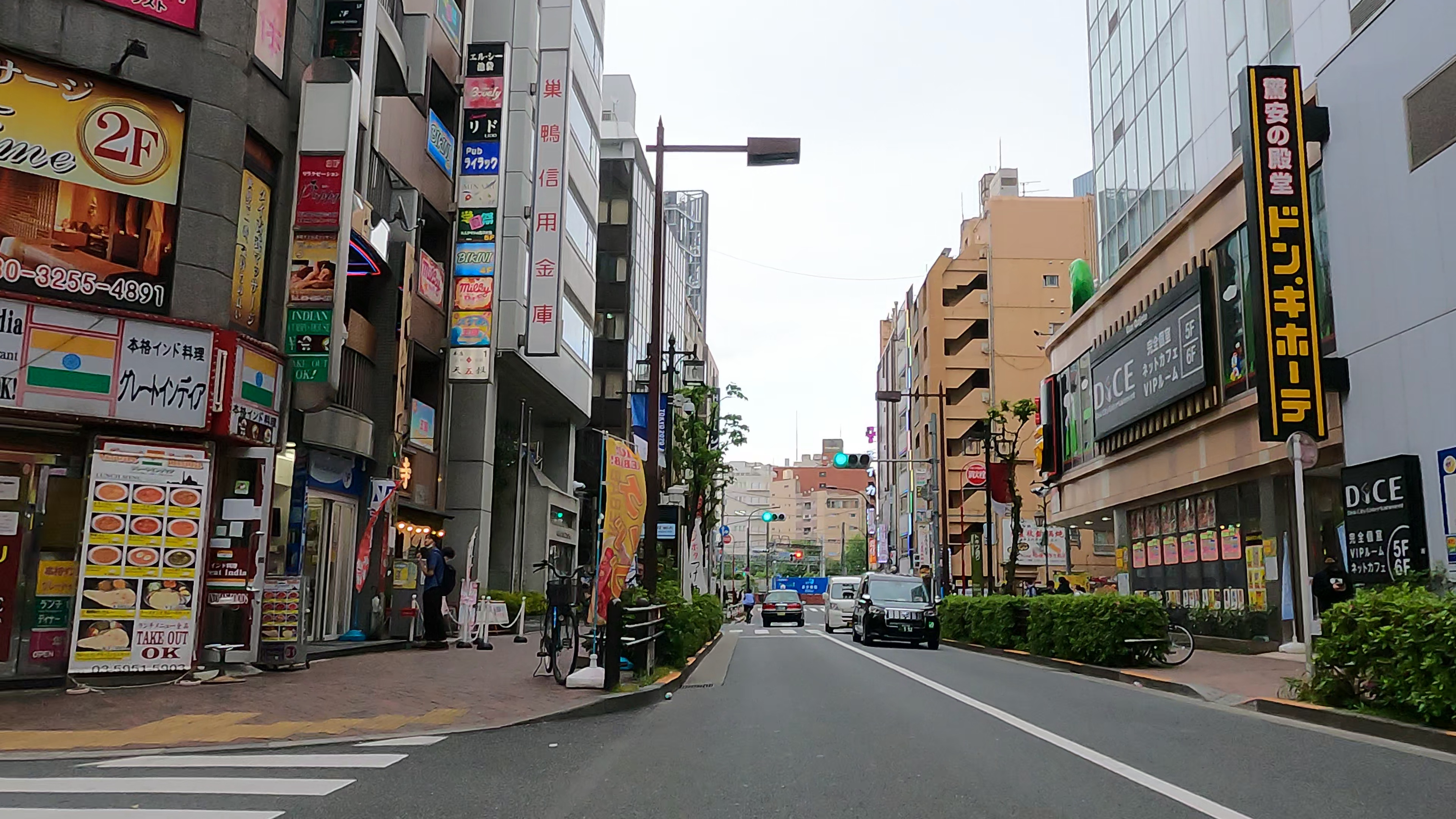}
      \end{minipage} \\
      
        \begin{minipage}[t]{0.3\linewidth}
        \centering
            \includegraphics[keepaspectratio,scale=0.234]{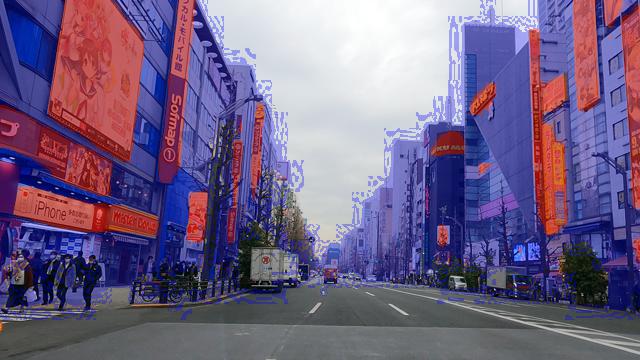}
            \caption{Akihabara}
            \label{img:akihabara}
      \end{minipage} &
      
        \begin{minipage}[t]{0.3\linewidth}
        \centering
            \includegraphics[keepaspectratio,scale=0.234]{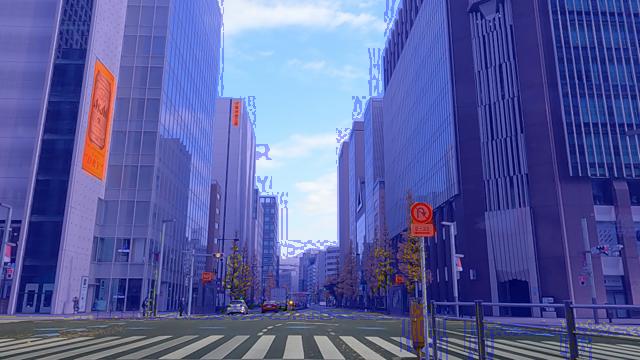}
            \caption{Marunouchi}
            \label{img:marunouchi}
      \end{minipage} &
      
        \begin{minipage}[t]{0.3\linewidth}
        \centering
            \includegraphics[keepaspectratio,scale=0.234]{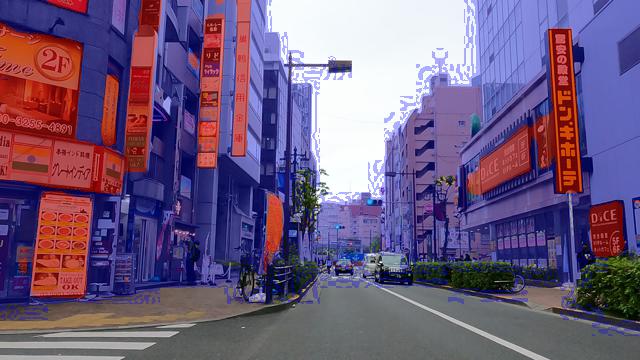}
            \caption{Ikebukuro}
            \label{img:ikebukuro}
      \end{minipage}
    \end{tabular}
  \end{figure*}

The developed models (DeepLab+Xc7 and DeepLabV3+ downscale) were implemented to infer the areas covered by the primary and secondary contour in three images of typical streetscapes in Tokyo. The upper row of Fig.\ref{img:akihabara} - Fig.\ref{img:ikebukuro} are the original images, the lower row illustrates the predictions of the models: the blue areas are those covered by the primary contour (building facades), and the red areas are those covered by the secondary contour (billboards). Visual inspection allows us to confirm the elevated accuracy of DeepLabV3+ downscale model (reported in the previous section) in recognizing the billboards in Japanese streetscapes.

Table \ref{tab:result} shows the quantified results of the predictions: the first and second rows report the ratio of the area covered by the primary and secondary contour over the entire image, and the third row is the ratio of the area covered by the secondary contour to that covered by the primary contour. The last quantity incorporates the aesthetic quality that Ashihara (1983) \cite{ashihara1983aesthetic} argued: the high ratio means the abundance of billboards in the streetscape, suggesting that such a streetscape may not be preferred by people in the aesthetic sense \cite{ashihara1983aesthetic}. Although this paper does not conclude the validity of such a claim owing to the absence of people's perception data on these streetscapes, these measures allow us to differentiate two resembling images (i.e., Fig. \ref{img:akihabara} and Fig.\ref{img:ikebukuro}).

\begin{table}[h]
\centering
\begin{tabular}{c|ccc}
                       & Akihabara & Marunouchi & Ikebukuro \\ \hline
Primary contour (\%) & 51.4\%    &      70.6\%         &    59.8\%    \\
Secondary contour (\%) & 16.5\%          & 1.6\%        &    17.6\%      \\
Ratio (S/P)            &    32.1\% & 2.3\%      &  29.4\%
\end{tabular}
\caption{Results of inferred areas covered by the primary and secondary contour, and their ratio.}
\label{tab:result}
\end{table}

\section{Discussion}
\subsection{Model accuracy}
Compared to the benchmark model, the developed models achieved higher accuracy of billboard segmentation. This improvement was enabled by (1) using a larger size of annotated data set of billboard images and (2) adding variations in data augmentation. With 63.17\% of IoU (DeepLabV3+ downscale), the quantification of the amount of visually recognizable billboards in Japanese streetscapes is now possible in a scalable manner.

Nevertheless, two major issues are present in the prediction accuracy. First, the presence of vegetation affected the segmentation of the primary contour: when a tree is present in front of the building facades, the corresponding area did not fall inside the primary contours in our models. This issue must be further discussed considering the aesthetic evaluation of street vegetation, as argued in Smardon (1988) \cite{smardon1988perception}. Second, the reflection of billboards on the glasses on buildings made the billboards segmentation less accurate. This problem could be mitigated by adding the amount of annotated data.

\subsection{Conceptual issues}
This study tried to operationalize the notion of the primary and secondary contour \cite{ashihara1983aesthetic} in the context of aesthetic evaluation of streetscapes. Because the original notion itself was not well defined, unable to attribute all the elements to either a constituent of the primary/secondary contour in a unique manner, we assumed that the secondary contour includes road signs and signs on glasses on building facades, in addition to billboards. This delineation may indeed be different from the original notion; however, the literature and this study have the same objective, which is to understand contributions of visually recognizable information on people's aesthetic evaluation on streetscapes. In this sense, our study prepared the first step to the understanding, with the aid of computational technologies.

\subsection{Future works}
\paragraph{Model development} the semantic segmentation models can be extended and their accuracy can be enhanced by adopting existing techniques. For instance, the adoption of transfer learning \cite{long2015fully}, in which the number of classes in the output layer is changed, is a possible option to build on DeepLabV3+ downscale model and the unique data set with billboard annotations.

Another possibility is using HRNet-OCR \cite{tao2020hierarchical} that achieved SoTA (state-of-the-art). At the cost of high computation costs (i.e., using multi GPU), it may further enhance the accuracy by introducing images of different resolutions in the hieratchical network. They can be trained with existing data sets such as Cityscapes \cite{cordts2016cityscapes} and Mapillary \cite{neuhold2017mapillary}, as well as our unique data set.

\paragraph{People's perception} empirically testing the theory of Ashihara (1983) \cite{ashihara1983aesthetic} requires another data set of people's perception and preference on streetscapes. A possible approach is to ask people to rate the degree of beauty of a given set of urban streetscape images. While it is a cost-effective method and allows to fix the perspective over different participants, the evaluation serves only as a proxy of people’s real perception of the streetscape on the place, because the images do not contain all the information that one may perceive on the place. The connection between the machinery evaluation of the streetscape and human perception should be argued in a future study.
%


%



\section*{Acknowledgment}
The authors would like to thank Mr. Taro Matsuzawa for the technical supports and annotators for their laborious works.

\section*{Data availability}
The unique data set of annotated images is available at \href{https://www.kaggle.com/yusukekumakoshi/billboard-in-japanese-streetscapes}{https://www.kaggle.com/yusukekumakoshi/billboard-in-japanese-streetscapes}. The developed model (DeepLabV3+ downscale) is available at \href{https://github.com/ursci/billboard-segmentation-model}{https://github.com/ursci/billboard-segmentation-model}.

\ifCLASSOPTIONcaptionsoff
  \newpage
\fi



\bibliographystyle{IEEEtran}
%

\bibliography{reference}




%








\end{document}